%% file: main.tex
\def\year{2021}\relax
\documentclass[letterpaper]{article} 
\usepackage{aaai21}  
\usepackage{times}  
\usepackage{helvet} 
\usepackage{courier}  
\usepackage[hyphens]{url}  
\usepackage{graphicx} 
\usepackage{natbib}  
\usepackage{caption} 
\usepackage{amsmath,amsfonts,amssymb}
\usepackage{booktabs}
\usepackage{makecell}
\usepackage{multirow}
\usepackage{tabularx}
\usepackage{textcomp}
\usepackage[switch]{lineno}  %

\mathchardef\mhyphen="2D

\newcommand{\ignorebig}[1]{}

\newcommand{\figref}[1]{Figure~\ref{#1}}

\usepackage{hyperref} 
\hypersetup{
    colorlinks=true,
    linkcolor=blue,
    filecolor=blue,      
    urlcolor=blue,
}

\title{Learning Object Detection from Captions via Textual Scene Attributes}
\author {
        Achiya Jerbi\textsuperscript{\rm 1},
        Roei Herzig\textsuperscript{\rm 1},
        Jonathan Berant\textsuperscript{\rm 1,\rm 4},
        Gal Chechik\textsuperscript{\rm 2,\rm 3},
        Amir Globerson\textsuperscript{\rm 1} \\
}
\affiliations {
    \textsuperscript{\rm 1} Tel Aviv University, 
    \textsuperscript{\rm 2} Bar-Ilan University 
    \textsuperscript{\rm 3} NVIDIA Research
    \textsuperscript{\rm 4} The Allen Institute for AI \\
}

\begin{document}
\maketitle

\begin{abstract}
\input{sections/abstract}
\end{abstract}

\input{sections/introduction}
\input{sections/related_work}

\input{sections/model}
\input{sections/experiments}
\input{sections/discussion}

\input{sections/acknowledgements}

\bibliography{papers_bib}

\end{document}

%% file: sections/abstract.tex
Object detection is a fundamental task in computer vision, requiring large annotated datasets that are difficult to collect, as annotators need to label objects and their bounding boxes. Thus, it is a significant challenge to use cheaper forms of supervision effectively. Recent work has begun to explore image captions as a source for weak supervision, but to date, in the context of object detection, captions have only been used to infer the categories of the objects in the image. In this work, we argue that captions contain much richer information about the image, including attributes of objects and their relations. Namely, the text represents a scene of the image, as described recently in the literature. We present a method that uses the attributes in this ``textual scene graph'' to train object detectors. We empirically demonstrate that the resulting model achieves state-of-the-art results on several challenging object detection datasets, outperforming recent approaches.

%% file: sections/introduction.tex
\section{Introduction} 
\label{introduction}

Object detection is one of the key tasks in computer vision. It requires detecting the bounding boxes of objects in a given image and identifying the category of each one. While object detection models have many real-life applications, they often also serve as a component in higher-level machine-vision systems such as Image Captioning, Visual Question Answering~\cite{anderson2018bottom}, Grounding of Referring Expressions ~\cite{hu2017modeling}, Scene Graph Generation~\cite{sg_generation_msg_pass,neural_motifs}, Densely Packed Scenes Detection~\cite{goldman2019dense},Video Understanding~\cite{zhou2019grounded,herzig2019spatio,materzynska2019something}, and many more.

The simplest way to train an object detection system is via supervised learning on a dataset that contains images along with annotated bounding boxes for objects and their correct visual categories. However, collecting such data is time consuming and costly, thus limiting the size of the resulting datasets. An alternative is to use weaker forms of supervision. The most common instance of this approach is the problem of \textit{Weakly Supervised Object Detection} (WSOD), where images are only annotated with the set of object labels that appear in them, but without annotated bounding boxes. Such datasets are of course easier to collect (e.g., from images collected from social media, paired with their user-provided hashtags \cite{mahajan2018exploring}), and thus much research has been devoted for designing methods that learn detection models in this setting. However, WSOD remains an open problem, as indicated by the large performance gap of $>$50\% between WSOD~\cite{singh2018sniper} and fully supervised detection approaches~\cite{ren2020instance} on the PASCAL VOC detection benchmark~\cite{everingham2010pascal}. 

\begin{figure}[t!]
	\begin{center}
        \includegraphics[width=\linewidth]{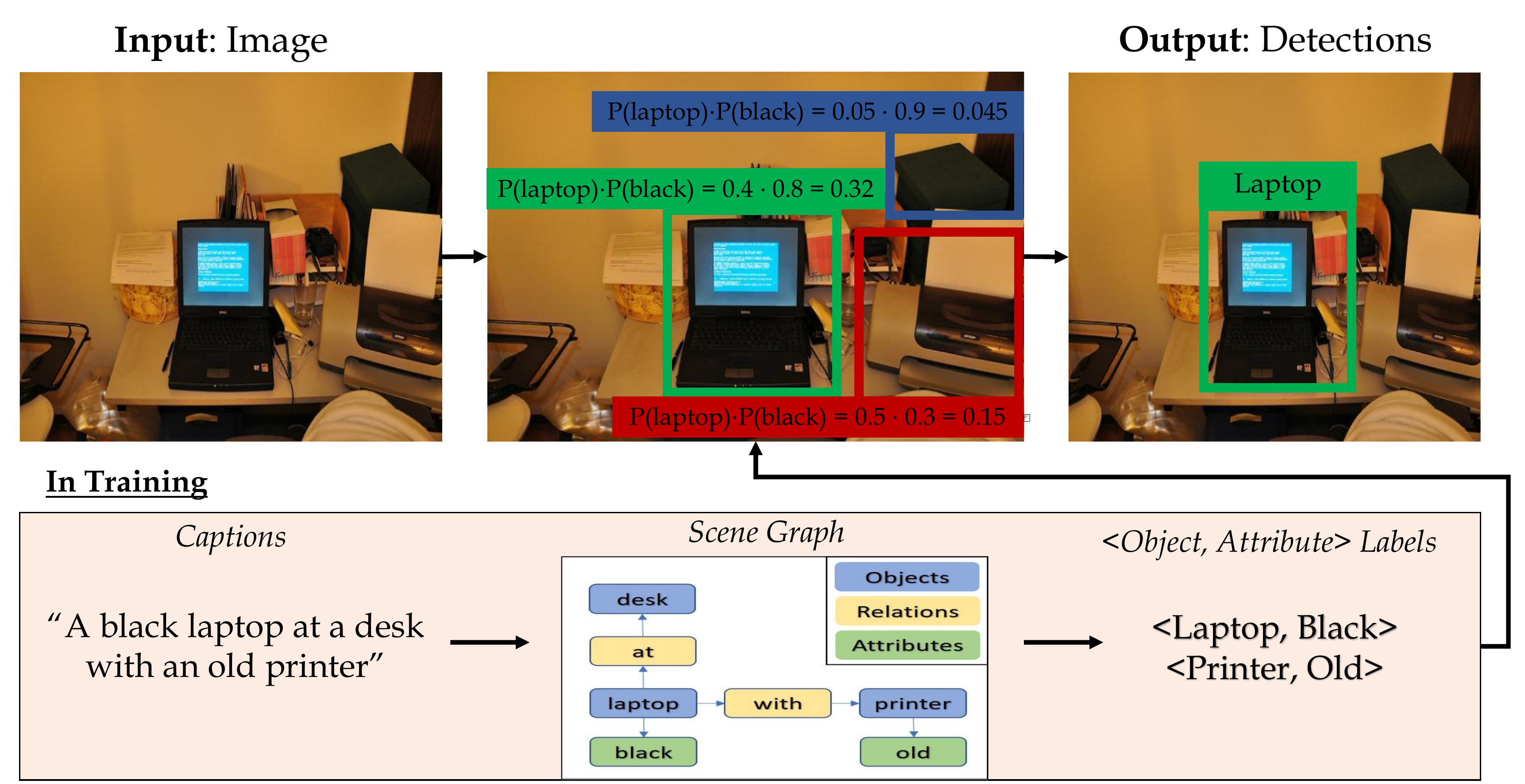}
	\caption{
    An illustration of our novel scene graph refinement process. The model makes use of the ``black'' attribute to localize the ``laptop'' object in the image at train time. This will result in improved object detection accuracy at test time.}
	\label{teaser}
	\end{center}
\end{figure}

An alternative, and potentially rich, source of weak supervision is image captions. Namely, textual descriptions of images, that are fairly easy to collect from the web. The potential of captions for learning detectors was recently highlighted in \citet{ye2019cap2det}, where they improve the extraction of object pseudo-labels from the captions. 

In this work, we argue that captions contain much richer information than has been used so far, because a caption tells us more about an image than just the objects it contains. It can reveal the attributes of the objects in the image (e.g., a blue hat) and their relations (e.g., dog on chair). In the machine vision literature, such a description of objects, their attributes and relations is known as a \emph{scene graph} (SG) \cite{johnson2015image}, and these have become central to many machine vision tasks \cite{img_retriev_using_sg, johnson2015image, sg_generation_msg_pass, support_relations,neural_motifs,herzig2018mapping}. This suggests that captions can be used to extract part of the scene graph of the image they accompany. 

Knowing the scene graph of an image provides valuable information that can used as weak supervision. To understand why, consider an image with two fruits that are hard to identify, and the caption ``a red apple next to a pear''. Since it is relatively easy to recognize a red object, we can use this knowledge to identify that the red fruit should have the ``apple'' label. An illustration of such possible usage of visual attributes in the object classification process from the COCO Captions dataset~\cite{chen2015microsoft} is shown in Figure \ref{teaser}.

We propose a learning procedure that captures this intuition by extracting ``textual scene graphs'' from captions, and using them within a weak supervision framework to learn object detectors. Our approach uses a novel notion of an entanglement loss that weakly constrains visual objects to have certain visual attributes corresponding to those describing them in the caption. Empirical evaluation shows that our model achieves significantly superior results over baseline models across multiple datasets and benchmarks.

Our contributions are thus: (1) We introduce a novel approach that aligns the structured representation of captions and images. (2) We propose a novel architecture with an entanglement loss that uses textual SGs to enforce constraints on the visual SG prediction. (3) We demonstrate our approach and architecture on the challenging MS COCO and Pascal VOC 2007 object detection datasets, leading to state of the art results and  outperforming prior methods by a significant gap.

%% file: sections/related_work.tex
\section{Related Work} 
\label{chap:Related_Work}

\textbf{Weakly Supervised Object Detection.} WSOD is a specific task out of a broader class of problems, named \textit{Multiple Instance Learning} (MIL). In MIL problems, instead of receiving a set of individually labeled instances, the learner receives a set of labeled bags, each containing many instances \cite{dietterich1997solving}. MIL is a valuable formalization for problems with a high input complexity and weak forms of supervision, such as medical images processing~\cite{quellec2017multiple}, action recognition in videos~\cite{ali2008human} and sound event detection~\cite{kumar2016audio}. The motivation behind the MIL formalization stems from the fact that the correct label for the input bag can be predicted from a single instance. For example, in the WSOD task, an object label can be inferred from the specific image patch in which the object appears. This is referred to as the standard multiple instance (SMI) assumption~\cite{amores2013multiple}, i.e., every positive bag contains at least one positive instance, while in every negative bag all of the instances are negative. 

Recent works as ~\citet{oquab2015object, zhou2016learning} use the MIL formalization and propose a new Global Max (or Average) Pooling layer to learn the activation maps for object classes. Moreover, ~\citet{bilen2016weakly} introduced Weakly-Supervised Deep Detection Networks (WSDDN) that use two different data streams for detection and classification, while the detection stream is used to weigh the classification predictions.

The works by \citet{akbari2019multi, gupta2020contrastive} aim to tackle the problem of \textit{Weakly Supervised Phrase Grounding}, which also requires finding an alignment between image regions and caption words. However, as their task objective is to find relevant image regions rather than to train an object detector, the image captions are given as inputs also at test time, and the task does not aim to correctly identify all of the existing objects in an image but only those that are present in its caption. Moreover, their task setting assumes the existence of a pretrained Faster-RCNN object detector for the region proposal extraction, which is not allowed in our setting.

Lately, the novel task of learning object detection directly from image captions was introduced by \citet{ye2019cap2det}. Their work addresses the same task as ours, but as they focus on achieving better object pseudo-labels from the image captions, we show that using the captions data more efficiently and providing the model with better image understanding abilities is a more important direction. Furthermore, 
\citet{ye2019cap2det} use additional supervision in the form of $<$image captions, object annotations$>$ pairs which is costly to collect, and here we show that our use of captions obviates the need for this additional supervision.

\subsubsection{Models for Images and Text.} Even though the problem of modeling the relationship between images and text has attracted many research efforts throughout the years, the task of training an object detector from image captions is relatively novel. Recently, there has been a surge of works trying to build a unified general-purpose framework to tackle tasks that involve both visual and textual inputs~\cite{lu2019vilbert,su2019vl,tan2019lxmert,chen2019uniter}. These works take large-scale parallel visual and textual data, pass it through an attention-based model to get contextualized representations, and apply a reconstruction loss to predict masked out data tokens (self-supervision). The resulting models were proven to achieve state-of-the-art results for various visual-linguistic tasks via transfer learning. This is related to recent advances in self-supervision in natural language processing, which train a transferable general-purpose language model with a similar masking objective~\cite{devlin2018bert}. However, while these works are useful for  scenarios that do not require prediction of the alignment between the visual and textual data, our objective is to explicitly classify the objects in the input image. Moreover, as we aim to train an object detector which naturally does not receive any textual input, we cannot use a model that requires both visual \emph{and} textual inputs.

\begin{figure*}[t!]
	\begin{center}
        \includegraphics[width=\linewidth]{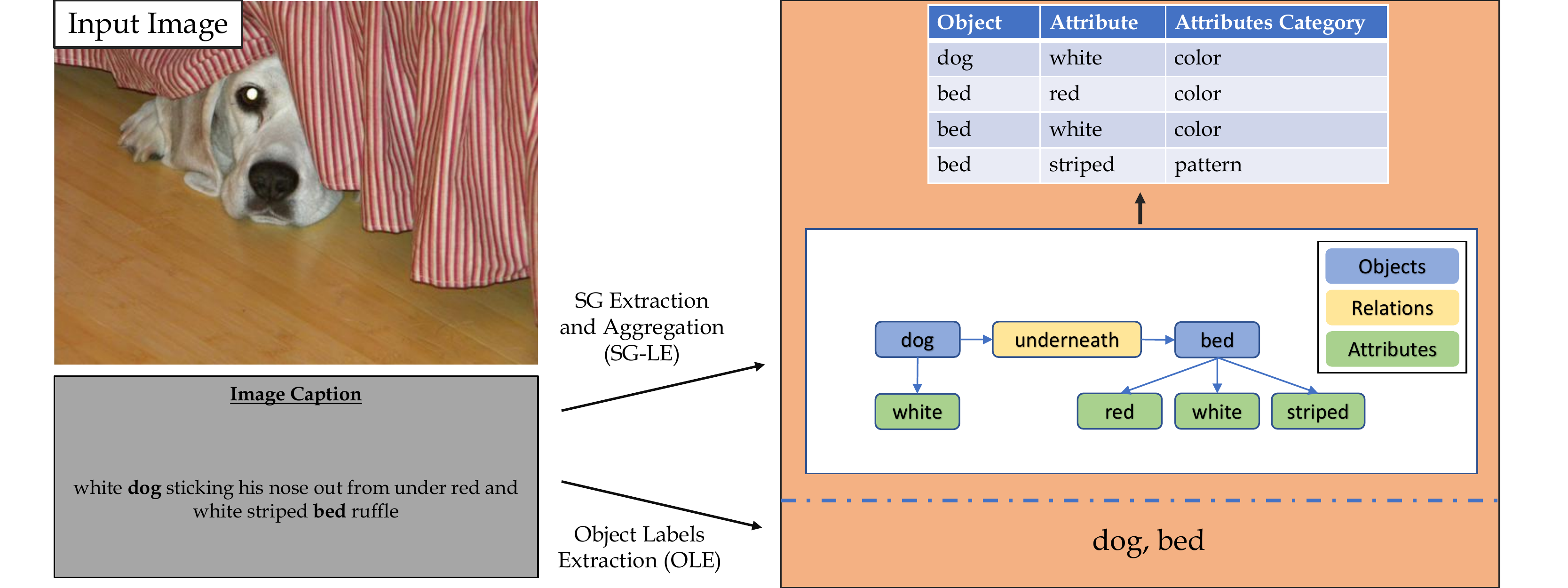}
	\caption{An illustration of our Textual Label Extraction (TLE) module. Given an image and its caption, we apply exact string matching to detect objects, and use a text-to-scene-graph model to generate a scene graph from the captions and aggregate object and attribute pairs.}
	\label{fig:label_extraction}
	\end{center}
\end{figure*}

\subsubsection{Scene Graphs.} The machine vision community has been using scene graphs of images for representation of information in visual scenes. SGs are used in various computer vision tasks including image retrieval~\cite{johnson2015image,schuster2015generating}, relationship modeling~\cite{referential_relationships,raboh2020dsg,Schroeder2019ICCV}, image captioning~\cite{xu2019scene}, image generation~\cite{herzig2019canonical,johnson2018image}, and recently even video generation~\cite{2020ActionGraphs}.

Textual scene graph prediction is the problem of predicting SG representations from image captions. While the problem of generating semantically meaningful directed graphs from a sentence is well known in NLP research as \textit{Dependency Parsing}, in the SG context the only three meaningful word classifications are objects, attributes and relations. Early approaches to this problem use a dependency parser as a basis for the SG prediction~\cite{schuster2015generating, wang2018scene}. Recently,~\citet{andrews2019scene} proposed to train a transformer model on a parallel dataset of image region descriptions and scene graphs, taken from the Visual Genome dataset~\cite{krishna2017visual}.

%% file: sections/model.tex
\section{The SG2Det Model}
\label{chap:sg2det}

We next describe our approach for using image captions to learn an object detector. We emphasize that image captions are our {\em only} source of supervision, and no manually annotated bounding boxes or even ground-truth image-level categories are used. We refer to our model as SG2Det since it uses textual scene graphs for learning object detectors. 

Image captions can provide rich and informative data about visual scenes. Namely, in addition to providing the categories of objects in the image, the captions can suggest the relations between the objects, their positions within the image and even their visual attributes. In this work, we claim that by aligning the scene graph structure as extracted from the captions to the different image regions, we can provide the model with an improved understanding of the visual scenes and obtain superior object detection results. 

The key element of our approach is an ``entanglement loss'' that aligns the visual attributes in the image with those described in the text. As an example, consider the image caption ``a red stop sign is glowing against the dark sky''. Instead of extracting only the ``stop sign'' object pseudo-label and discarding the rest of the caption information, we propose to use the ``red'' attribute to enhance our supervision. Namely, instead of training the model to find the image region that is the most probable to be a ``stop sign'', our training objective is now to find the one that is the most probable to be both a ``stop sign'' and ``red''. Technically, this is achieved by multiplying object and attribute probabilities, as described later in the The Attribute Entanglement Loss subsection.

Our proposed SG2Det model is, therefore, composed of the following three components:

\begin{enumerate}
    \item The \textit{Textual Label Extraction} (TLE) module, which extracts the object pseudo-labels and the scene graph information from the text captions.
    \item The \textit{Visual Scores Extraction} (VSE) module, which finds bounding box proposals and outputs logits for the categories and attributes of these boxes.
    \item The \textit{Attribute Entanglement Loss} (AEL) module, which enforces agreement between the textual and visual representations.
\end{enumerate}

\begin{figure*}[t!]
	\begin{center}
        \includegraphics[width=\linewidth]{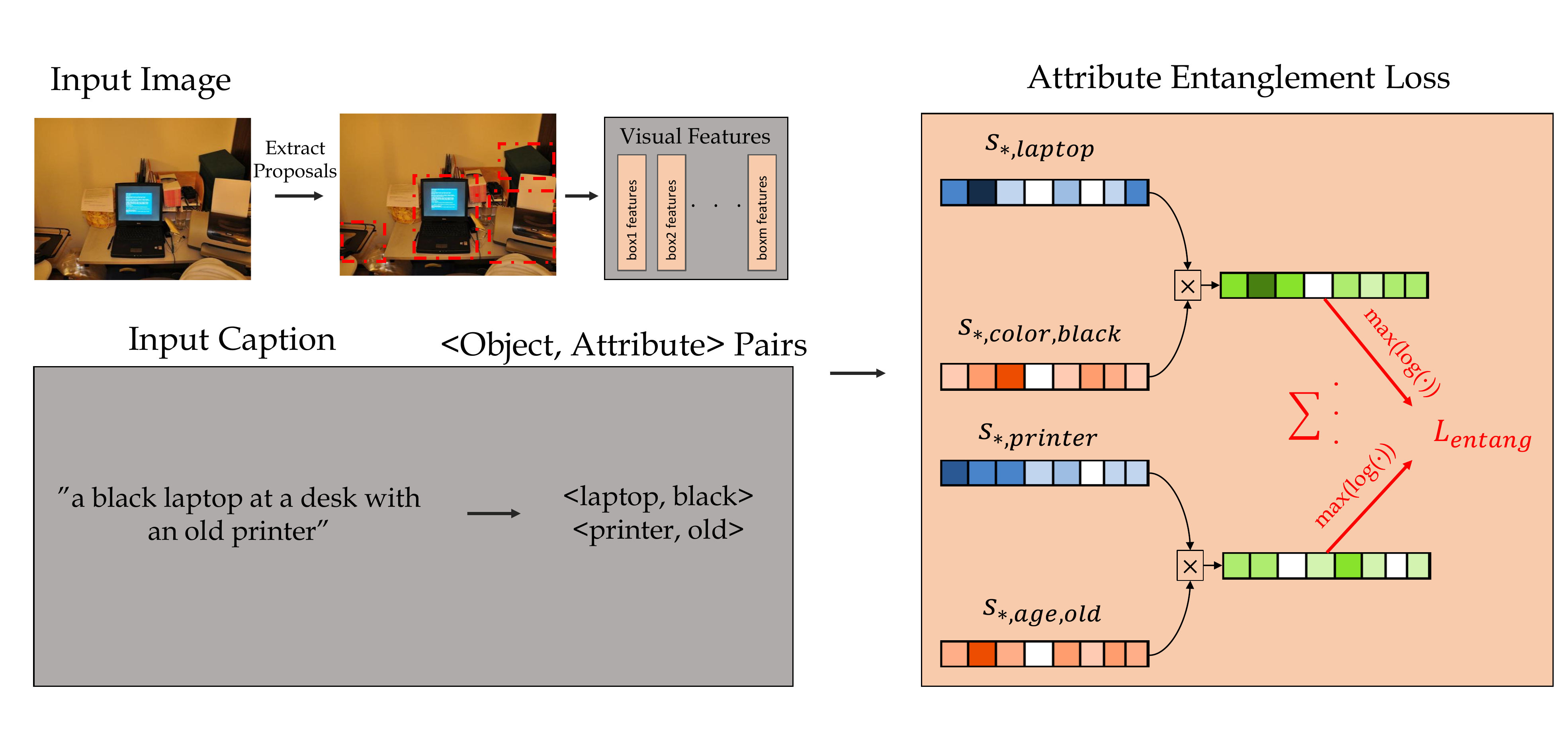}
	\caption{An overview of our attribute entanglement loss. First, region proposals (bounding boxes) are extracted from the image, and convolutional features are calculated for each. The image captions are used to extract a list of object and attribute pairs. 
	The scores $s_{i,c},s_{i,a,v}$ are calculated for the object categories and attributes, and they are used within the entanglement loss that captures the product scores of the object-attribute pairs.
	}
	\label{sg_example}
	\end{center}
\end{figure*}

\subsection{Textual Label Extraction}
\label{sg2det:label_extraction}
In this module, we extract the object pseudo-labels and the scene graph data from the image captions. ~\figref{fig:label_extraction} provides a high-level illustration of this module. The object category labels can be extracted in several different ways (e.g. string matching, synonym dictionary, a trained text classifier, etc.) as explored by \citet{ye2019cap2det}, but we choose simple string matching to highlight the usefulness of our novel loss. 
In addition to the object category labels, we also extract a textual scene graph representation for each of the captions, and use the $<$object, attribute$>$ pairs aggregated from them within the entanglement loss described later. For SG extraction, we use an off-the-shelf textual scene graph parser\footnote{\url{https://github.com/vacancy/SceneGraphParser}} based on~\citet{schuster2015generating}. We also experimented with scene graphs extracted by ~\citet{wang2018scene}, but found these to achieve inferior results.

We choose to split the object attributes to categories (e.g., color, shape, material, etc.) based on a categorization taken from the GQA dataset~\cite{hudson2019gqa}. This dataset contains visual questions that are automatically generated using scene graphs. By looking at the semantic representation (logical form) of the questions, we can derive the $22$ different attribute categories that were used by the authors for dataset generation. Some of these attribute categories are general (e.g. \texttt{color}, \texttt{size}), while others are specific to certain object classes (e.g. \texttt{shape}, \texttt{pose}, \texttt{sportActivity}).

The advantage of this categorization is that it provides the model with additional knowledge. Unlike object labels, each region proposal can have more than one attribute. For example, a cat can be both \emph{large} and \emph{brown}. Thus, we cannot model the attributes prediction as a multi-class problem. By using categories, we enforce mutual exclusivity within each category, preventing the model, for example, from predicting that an object is both \emph{black} and \emph{white}; while allowing multiple labels across different categories.

The output of this stage for each image is as follows:
\begin{itemize}
    \item A set $O$ of object categories. For example $O=\{\emph{cat},\emph{dog}\}$ indicates that the text describes a cat and a dog. 
    \item For each $o\in O$ we have a set $A_o$ containing attribute-value pairs $(a,v) \in A_o$. Thus, $A_\textit{cat}=\{(\texttt{color},\textit{brown}),(\texttt{size},\textit{large})\}$ indicates that the cat is brown and large.
\end{itemize}

\subsection{Visual Scores Extraction}
\label{sg2det:visual_scores_extraction}

We now consider the input image and extract bounding boxes from it. Then, fully connected (FC) classifiers are applied to these bounding boxes, resulting in model scores for object categories and attributes. Below we elaborate on this process.

First, we generate object region proposals $r_1, \ldots, r_m$ using the \textit{Selective Search} algorithm ~\cite{uijlings2013selective}, and compute a convolutional feature map for each input image by feeding it to a convolutional network pre-trained on ImageNet \cite{deng2009imagenet}. Importantly, the convolutional backbone is not trained on an object detection dataset since our objective is to learn detection from the captions only.  Then, we apply a ROIAlign layer~\cite{he2017mask} for cropping the proposals to fixed-sized convolutional feature maps. Finally, a box feature extractor is applied to extract a fixed-length descriptor $\phi(r_i)$ for each proposal $r_i$.

Denote by $C$ the number of different object classes and by $m$ the number of different regions. We now apply an FC classifier followed by a softmax layer to achieve an object score $s_{i,c}$ for each $i\in\{1,\ldots,m\}$ and $c \in \{1, \ldots, C+1\}$, where $C+1$ is the background class. Similarly, for each region $i$, attribute type $a$ and attribute value $v$, we obtain a score $s_{i,a,v}$ via an FC prediction head for attribute $a$. 

The output of this stage is as follows:
\begin{itemize}
    \item A score $s_{i,c}$ for each bounding box (region) $i$ and category value $c$.
    \item A score $s_{i,a,v}$ for each bounding box (region) $i$, attribute type $a$ and attribute value $v$. E.g., $a$ can be ``shape'' and $v$ can be ``rectangular''.
\end{itemize}

\subsection{The Attribute Entanglement Loss}
\label{sg2det:attribute_entanglement_loss}

The core element of our approach is a loss that enforces agreement between the textual and visual representations of the image. To achieve this, we adapt the MIL approach to capture also the attributes information.

We begin by describing the standard loss used in MIL for the case where only object categories information is available in both the textual and visual descriptions. Namely, assume that from the text side we only have the set of object categories $O$, and from the image side we only have the model scores for all categories and bounding boxes $s_{i,c}$. The intuition in this case is that each category in $O$ should have at least one bounding box describing it. Thus, if $\emph{cat}\in O$ then $s_{i,\emph{cat}}$ should be high for some $i$. This motivates the use of the following loss:
\begin{equation}
    L_{\text{obj}} = -\frac{1}{|O|} \sum_{c\in O} \max_i\{\log{s_{i,c}}\}
    \label{eq:l_obj}
\end{equation}

 \begin{figure*}[ht!]
	\begin{center}
        \includegraphics[width=\linewidth]{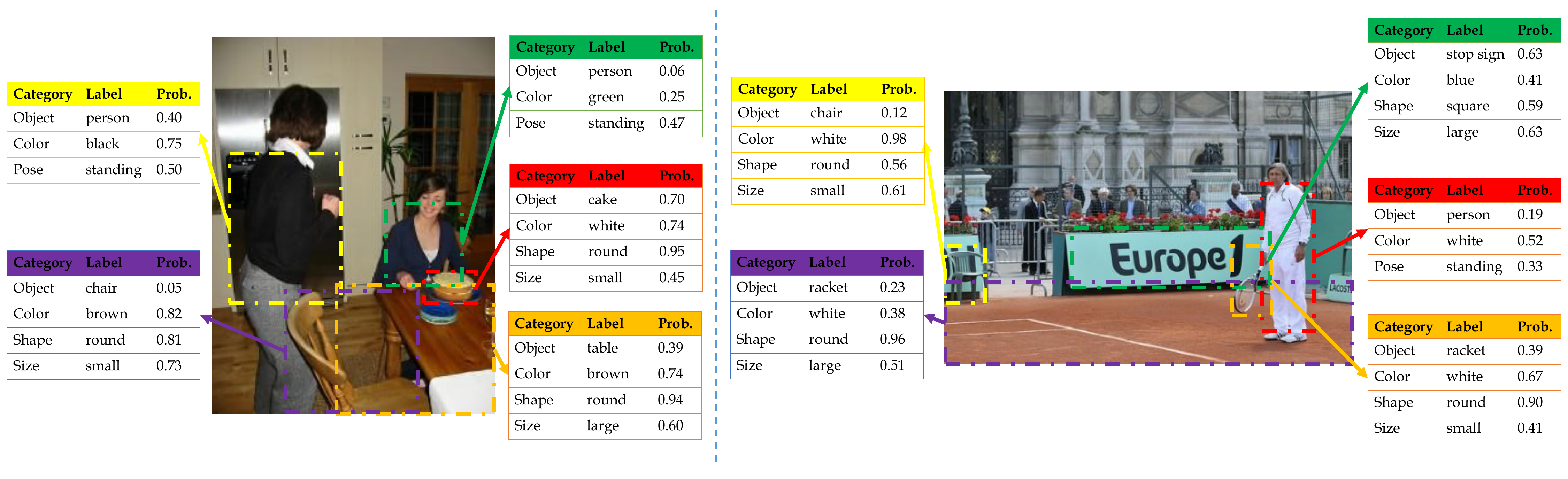}
	\caption{Qualitative examples for our objects and attributes predictions. It can be seen that our attribute classifiers provides meaningful classification results across different categories, which validates the success of our attribute classification objective.}
	\label{sg_qualitative}
	\end{center}
\end{figure*}

However, in our case we wish to go beyond object information and use attributes. Namely, if we have $\emph{cat}\in O$ and $(\texttt{color},\textit{brown})\in A_{\text{cat}}$, we would like some bounding box to {\em both} contain a cat and have the attribute brown. Namely, there should be a box $i$ where both $s_{i,\textit{cat}}$ {\em and} $s_{i,\texttt{color},\textit{brown}}$ are high. The use of {\em and} here is important, since a violation of either these conditions would imply this box does not contain the brown cat. The following entanglement loss precisely captures the intuition that attributes and categories should be dependent:
\begin{equation}
    L_{\text{entang}} = -\frac{1}{|O|}\sum_{c\in O, (a,v)\in A_c} \max_i\{\log\left(s_{i,c} \cdot s_{i,a,v}\right)\}
    \label{eq:l_entang}
\end{equation}
We note that although the log in the above can be written as $\log\left(s_{i,c}\right)+\log \left(s_{i,a,v}\right)$, this objective is very different from using two objectives like \eqref{eq:l_obj}, one for attributes and one for objects. This is because the maximum over $i$ is applied to the log, and thus if one of $s_{i,c}$ or $s_{i,a,v}$ is very low, the bounding box $i$ will not be the maximizer.

We conclude by emphasizing that the whole training procedure operates without any explicit supervision of bounding boxes. Despite this, we shall see that the model succeeds in learning both detection (i.e., finding the right bounding boxes) as well as classification for both object categories and attributes.

\subsection{Other Losses}
In addition to the losses ${L}_{\text{entang}}$ and ${L}_{\text{obj}}$, we use a loss that promotes high scores for categories in $O$ and low scores for categories outside $O$. This is referred to by \citet{tang2017multiple} as the \textit{Multiple Instance Detection} (MID) loss. We now pass our region descriptors through two parallel FC layers to get two different $(m \times C)$-dimensional matrices. Then, we pass one of the matrices through a sigmoid function and the other through a softmax layer over the different regions, and multiply them element-wise to get a score for each $<$region, object label$>$ pair. We denote these scores by $s^{\text{mid}}_{i,c}$ for each $i\in\{1,\ldots,m\}$ and $c \in \{1, \ldots, C\}$. Note that we do not add a ``background'' class here since the MID loss cannot propagate gradients for this class. Next, we aggregate the scores $s^{\text{mid}}_{i,c}$ from all different boxes to a single image-level soft-binary score $\hat{y}_c\in[0,1]$ as follows:
\begin{equation}
    \hat{y}_c = \sigma(\sum_{i=1}^{m} s^{\text{mid}}_{i,c})
\end{equation}
where $\sigma$ is the sigmoid function. Now, we consider the binary cross entropy between $\hat{y}_c$ and the indicator corresponding to $O$. This results in the following loss term:
\begin{equation}
    L_{\text{mid}} = -\sum_{c=1}^{C} I[c\in O]\log{ \hat{y}_c} + I[c\notin O]\log\left(1-\hat{y}_c\right)
\label{eq:mid}
\end{equation}

Our overall loss is thus:
\begin{equation}
     L_{\text{total}} =  L_{\text{mid}} + \lambda_1 L_{\text{obj}} + \lambda_2 L_{\text{entang}}
\end{equation}

\subsection{Online Refinement}
So far we assumed that for each category $c\in O$ only one bounding box contains the object pseudo-label (namely, the one that maximizes $s_{i,c}$). In practice, there could be other bounding boxes that highly overlap with the maximizing one, and should therefore be included in the loss. This intuition was used by the \textit{Online Instance Classifier Refinement} (OICR) method introduced in ~\citet{tang2017multiple}, and also 
in \citet{ye2019cap2det}. In order to provide a fair comparison to \citet{ye2019cap2det} we also use OICR here. 

The OICR method uses $K$ different score functions $s^{k}_{i,c}$ for $k\in\{1,\ldots,K\}$ and $L^k$ corresponding losses. The loss $L^k$ is similar to Eq. \eqref{eq:l_obj} but with two differences: it uses all boxes that sufficiently overlap the maximizing box, and the $\max$ operator is applied to the scores from the previous OICR step. This provides a refinement process where each FC classifier uses the output of its predecessor as an instance-level supervision. The first score function $s^{0}_{i,c}$ is obtained by applying a softmax function over the $s^{\text{mid}}_{i,c}$ scores from the MID step.
Here, we extend OICR to also use attribute scores $s^{k}_{i,a,v}$ and similarly extend the loss in Eq. \eqref{eq:l_entang}.

Note that we do not apply an MID loss to the attributes as we do for objects (i.e., aggregating attribute scores from all regions, and comparing to the attribute pseudo-labels extracted from the captions) since we found this to harm the performance of our model. We hypothesize that this is because the assumption that a label is present in the caption \textit{if and only if} it is present in the image is improbable for attributes, as image captions data is often sparse with attribute annotations. Thus, we do not have initial MID scores for the attributes as we do for the objects. Because of that, at the first OICR iteration we only train the attributes classifiers using the objects MID scores, and we do not apply the entanglement loss. E.g., if the image contains a ``brown cat'', we use the box that maximizes the ``cat'' probability as a supervision for the color classifier with the label ``brown''.

%% file: sections/experiments.tex
\section{Experiments}
\label{sec:experiments}

In this section, we show both qualitative and quantitative result for our SG2Det model. We show that compared to prior work, the additional scene graph information we extract from the image captions is indeed helpful, and provides significantly better object detection results on all of the benchmarks we evaluate on, without using any additional training data. Our model achieves state-of-the-art results on the COCO detection test-dev set and the PASCAL VOC 2007 detection test set, when training on multiple captions datasets. Specifically, when training on COCO captions, we achieve results that are comparable to the state-of-the-art on the PASCAL VOC 2007 test set for a WSOD model \textbf{that was trained on COCO ground-truth labels}.

\subsection{Implementation Details}
\label{experiments:implementation_details}

To the best of our knowledge, the only existing work to tackle the problem of training an object detector from image captions is \citet{ye2019cap2det}. Therefore, to ensure a fair comparison between our works, we use the same algorithm and configuration for proposal boxes extraction and the same convolutional backbone and feature layers, and our model is based on their official paper implementation.\footnote{\url{https://github.com/yekeren/Cap2Det}} Specifically, we use the Selective Search algorithm \cite{van2011segmentation} to extract (at most) 500 proposals for each image, taken from the OpenCV library. 

We compute the region descriptor vectors by using the (“Conv2d1a7x7” to “Mixed4e”) layers from InceptionV2~\cite{szegedy2016rethinking} for extracting the convolutional feature maps from the images. In addition, we use the (“Mixed5a” to “Mixed5c”) layers in the same model to extract the region descriptors after the ROIAlign \cite{he2017mask} operation. Finally, the convolutional backbone network was pre-trained on ImageNet \cite{deng2009imagenet}. 

\input{tables/voc_results}
\input{tables/coco_testdev_results}

We use the AdaGrad optimizer with a learning rate of 0.01 and set the batch size to 2. For the training data augmentation, we randomly flip the image left to right at training time and resize each image randomly to one of the four scales $s \in \{400,600,800,1200\}$. We set the number of OICR iterations to $K=3$, as we found this to yield the best performance. We use non-maximum-suppression (NMS) at the post-processing stage with an intersection-over-union (IOU) threshold of 0.4. We Follow \citet{ye2019cap2det} by weighing the $L_{\text{obj}}$ term by $\lambda_1 = 0.5$. For the $L_{\text{entang}}$ term we experiment with different values in \{$1e-2, 3e-3, 1e-3$\} and find $\lambda_2 = 1e-2$ to yield the best results.

Unlike \citet{ye2019cap2det}, we found that the performance of our SG2Det model continues to improve until 1M training steps when training on COCO Captions ($\sim$17 epochs) and 300K steps for Flickr30K ($\sim$19 epochs). We hypothesize that this is a result of the more complex optimization objective of our model. We pick the best model based on the validation set for each dataset. Our models are trained on a single Titan XP GPU for 7 days when training on COCO, and 2 days when training on Flickr30K.

As we experienced instability in the results when training the same model different times with different random seeds, all of the caption models' results we report were achieved by training the model 3 times, while the best one is chosen based on the validation set. Therefore, for almost all of the baseline models we report improved results over what is reported by \citet{ye2019cap2det}, when some of the models have significant improvement gaps.

\subsection{Datasets}
\label{experiments:datasets}

For training the SG2Det model, we use two popular image captions datasets: COCO Captions~\cite{chen2015microsoft} and Flickr30K~\cite{young2014image}. For training on the COCO Captions dataset, we use $118,287$ images, each paired with five different human-generated captions, which sum up to $591,435$ captions in total. For training on the Flickr30K dataset, we use $31,783$ images, each paired with five different human-generated captions, which sum up to $158,915$ captions in total. When evaluating the VOC 2007 dataset, we use the train and validation set for validation and report our results on the test set. As the object labels vocabulary differs between the COCO and VOC datasets, when evaluating on VOC we use only the twenty overlapping VOC objects. We report the mAP@0.5 for each of the object labels and its mean across the labels for this dataset.  When evaluating on COCO, we use the val2017 set for validation and test our model by submitting our object detection predictions to the COCO test-dev2017 evaluation server.\footnote{\url{https://competitions.codalab.org/competitions/20794}} We report the metrics provided by the server, where mAP@0.5:0.95  is the primary evaluation metric of the dataset.

\subsection{Models}
\label{experiments:models}

On all different benchmarks, we report our results for the following three models:

\begin{enumerate}
  \setlength{\itemsep}{0pt}
  \setlength{\parskip}{0pt}
  \setlength{\parsep}{0pt}
  \item The \textsc{Exact Match (EM)} baseline model proposed by \citet{ye2019cap2det}. This model performs a simple string matching to extract object pseudo-labels from the captions.
  \item The \textsc{EM + TextClsf} model. This is the best model reported by \citet{ye2019cap2det}. This model performs object pseudo-label extraction by training a text classifier on \emph{additional data} of parallel image captions and object annotation pairs, taken from the COCO detection dataset.
  \item Our novel \textsc{EM + SG Loss} model. Our model also performs exact string matching for object label extraction, but additionally applies our novel SG entanglement loss. \textbf{Note that our model is identical to \textsc{ExactMatch} when weighting the SG entanglement loss by 0.}
\end{enumerate}

\subsection{Results}
\label{experiments:results}

Figure \ref{sg_qualitative} shows qualitative examples for predictions of our SG2Det model. For consistency, we only visualize objects with detection confidence $> 5\%$, and attribute categories that are meaningful across different object classes. We can see that the model identifies most of the objects and their attributes correctly. This further validates our claim that our SG loss gives the model better scene understanding abilities, which in turn allow it to obtain improved object detection results. This figure also shows the quality of the attribute classifiers we obtain as a by-product of our training process.

Table~\ref{tab:result_cap_pascal_map} shows the results of our models on the VOC 2007 test set. At the top of the table, we show the results of models trained using the gold objects annotations on the COCO and VOC datasets (without bounding boxes annotations), as reported by~\citet{ye2019cap2det}. These results can be viewed as an upper bound for what is achievable by training from image captions using our method, since we use a weaker form of supervision.

In the middle part of the table, we report the VOC 2007 test results for models that were trained on COCO captions. Our novel \textsc{EM + SG Loss} model achieves state-of-the-art results on this dataset. It is worth noting that the performance of our model that was trained only on COCO Captions ($45.9$ mAP score), achieves comparable results to the WSOD model trained on the ground-truth COCO labels ($46.3$). This implies that our SG loss utilizes the image captions with close-to-maximal efficiency.

At the bottom of the table, we report the VOC 2007 test results for models trained on captions from the Flickr30K dataset. As before, our model achieves state-of-the-art results, which validates the contribution of our novel SG loss.

Table \ref{tab:result_cap_coco_map} shows the results of our model on the COCO testdev dataset. The main metric for evaluation for this dataset is mAP @ 0.5:0.95, which is reported in the leftmost column. Our novel \textsc{EM + SG Loss} model achieves state-of-the-art results on this baseline too.

To summarize, both our work and \citet{ye2019cap2det} seek to improve over the simple \textsc{ExactMatch} baseline. Our method uses contextual scene understanding for this purpose, while the \textsc{EM + TextClsf} method focuses on achieving better object-pseudo labels. When training on COCO Captions and predicting on VOC we can see that our model almost \textbf{doubles the performance gap} over the text classifier model, and when training on Flickr30K and evaluating on VOC, or training on COCO Captions and evaluating on COCO test-dev, our model's performance gap is about \textbf{4 times} the text classifier gap, without using any additional training data. This validates our hypothesis that better scene understanding is a more critical factor for WSOD models than extraction of better pseudo-labels.

%% file: tables/voc_results.tex
\begin{table*}[ht!]
    \footnotesize
    \centering
    \setlength\tabcolsep{0pt} 
    \begin{tabularx}{\textwidth}{p{3.0cm}|*{20}{>{\centering\arraybackslash}X}|>{\centering\arraybackslash}X}
    \toprule
        Methods & \rotatebox{90}{aero} & \rotatebox{90}{bike} & \rotatebox{90}{bird} & \rotatebox{90}{boat} & \rotatebox{90}{bottle} & \rotatebox{90}{bus} & \rotatebox{90}{car} & \rotatebox{90}{cat} & \rotatebox{90}{chair} & \rotatebox{90}{cow} & \rotatebox{90}{table} & \rotatebox{90}{dog} & \rotatebox{90}{horse} & \rotatebox{90}{mbike} & \rotatebox{90}{person} & \rotatebox{90}{plant} & \rotatebox{90}{sheep} & \rotatebox{90}{sofa} & \rotatebox{90}{train} & \rotatebox{90}{tv} & \rotatebox{90}{mean} \\
    \Xhline{2\arrayrulewidth}
        \multicolumn{21}{l}{Training on different datasets using ground-truth labels:} \rule{0pt}{1.01em} & \\
        \textsc{GT-Label VOC} & 68.7 & 49.7 & 53.3 & 27.6 & 14.1 & 64.3 & 58.1 & 76.0 & 23.6 & 59.8 & 50.7 & 57.4 & 48.1 & 63.0 & 15.5 & 18.4 & 49.7 & 55.0 & 48.4 & 67.8 & 48.5 \\ 
        \textsc{GT-Label COCO} & 65.3 & 50.3 & 53.2 & 25.3 & 16.2 & 68.0 & 54.8 & 65.5 & 20.7 & 62.5 & 51.6 & 45.6 & 48.6 & 62.3 & 7.2 & 24.6 & 49.6 & 34.6 & 51.1 & 69.3 & 46.3 \\
    \hline
        \multicolumn{21}{l}{Training on COCO dataset using captions:} \rule{0pt}{1.01em} & \\
        \textsc{ExactMatch (EM)} & 62.0 & 45.5 & 52.8 & 31.5 & 14.2 & 66.8 & 50.7 & 34.0 & 12.3 & 53.4 & 51.9 & 57.9 & 45.4 & 59.7 & 11.2 & 13.1 & 57.8 & 46.1 & 50.0 & 53.8 & 43.5 \\ 
        \textsc{EM + TextClsf} & 64.1 & 46.7 & 47.9 & 31.6 & 12.4 & 70.0 & 53.0 & 56.5 & 17.9 & 60.5 & 37.6 & 59.4 & 47.4 & 59.4 & 25.0 & 0.2 & 44.4 & 49.8 & 43.3 & 55.8 & 44.1 \\
        \textsc{EM + SG Loss} & 61.9 & 48.3 & 53.5 & 32.3 & 15.8 & 65.7 & 50.3 & 54.2 & 16.0 & 61.1 & 48.9 & 68.0 & 49.1 & 57.1 & 15.5 & 16.0 & 52.7 & 56.1 & 46.0 & 49.2 & \textbf{45.9}\\
    \hline
        \multicolumn{21}{l}{Training on Flickr30K dataset using captions:} \rule{0pt}{1.01em} & \\
        \textsc{ExactMatch (EM)} & 43.5 & 27.3 & 41.7 & 15.2 & 9.0 & 31.9 & 47.7 & 67.2 & 11.0 & 45.6 & 28.9 & 65.6 & 28.9 & 51.8 & 31.3 & 5.6 & 34.7 & 33.7 & 23.6 & 46.0 & 33.1 \\
        \textsc{EM + TextClsf} & 37.3 & 35.6 & 46.0 & 18.9 & 9.8 & 45.6 & 45.4 & 57.3 & 15.1 & 46.1 & 19.4 & 67.5 & 36.0 & 52.2 & 8.2 & 0.1 & 46.5 & 30.9 & 27.8 & 46.1 & 34.6 \\
        \textsc{EM + SG Loss} & 50.1 & 45.0 & 46.8 & 12.5 & 10.0 & 40.8 & 44.9 & 61.2 & 15.4 & 42.9 & 43.3 & 69.3 & 32.1 & 43.5 & 3.5 & 3.4 & 37.0 & 42.1 & 27.75 & 48.0 & \textbf{36.0} \\
    \bottomrule
    \end{tabularx}
    \vspace{-0.2cm}
    \caption{\textbf{Average precision on the VOC 2007 test set (learning from ground-truth annotations, COCO and Flickr30K captions)}. We train the detection models using the 80 COCO objects, but evaluate on only the overlapping 20 VOC objects.}
    \label{tab:result_cap_pascal_map}
\end{table*}

%% file: tables/coco_testdev_results.tex
\begin{table}[ht!]
    \footnotesize
    \centering
    \setlength\tabcolsep{0pt} 
    \begin{tabularx}{\linewidth}{>{\hsize=2.5\hsize}X|*{3}{>{\hsize=0.75\hsize\centering\arraybackslash}X}|*{3}{>{\hsize=0.75\hsize\centering\arraybackslash}X}}
    \toprule
        \multirow{2}{*}{Methods} & \multicolumn{3}{c|}{Avg. Precision, IoU} & \multicolumn{3}{c}{Avg. Precision, Area} \\
        & 0.5:0.95 & 0.5 & 0.75 & S & M & L \\
    \midrule
        \textsc{GT-Label} & 10.6 & 23.4 & 8.7 & 3.2 & 12.1 & 18.1 \\
    \hline
        \textsc{ExactMatch (EM)} \rule{0px}{1.01em} & 9.0 & 20.2 & 7.1 & \textbf{2.4} & 10.6 & 16.3 \\
        \textsc{EM + TextClsf}& 9.1 & 20.0 & \textbf{7.4} & 2.1 & 10.5 & 16.5 \\
        \textsc{EM + SG Loss} & \textbf{9.4} & \textbf{21.1} & 7.3 & \textbf{2.4} & \textbf{10.8} & \textbf{17.6} \\

    \bottomrule
    \end{tabularx}
    \caption{COCO test-dev mean average precision results when training on COCO captions. These numbers are computed by submitting our detection results to the COCO evaluation server. The best method is shown in \textbf{bold}.}
    \label{tab:result_cap_coco_map}
\vspace{-0.5cm}
\end{table}

%% file: sections/discussion.tex
\section{Discussion}
\label{sec:discussion}
We present a novel weakly supervised object detection approach that uses only image captions as supervision. Unlike previous approaches to this problem, we make use of the rich information available in the text in the form of visual attribute descriptions. We propose a novel entanglement loss that captures the coupling between the objects and attributes.

Our evaluation of the COCO and VOC datasets demonstrates state-of-the-art results, using less supervision than previous caption-based methods used by \citet{ye2019cap2det}. Moreover, it shows the power of using grounding information from the text when analyzing images. Here our focus was on attributes only, although textual scene relations can also be explored. These are more technically challenging to handle since the number of potential relations is quadratic in the number of region proposals. However, these can be pruned in different ways. Finally, here we used a fixed pre-trained text-to-scene-graph model. An exciting question is how to learn this jointly with the detection model. We leave these directions for future work.

%% file: sections/acknowledgements.tex
\section{Acknowledgements} 
\label{acknowledgements}
This work was supported by the Israeli Innovation Authority MAGNETON program, and the Israel Science Foundation.